\documentclass{article}

\PassOptionsToPackage{authoryear}{natbib}
 \usepackage[position, final]{neurips_2026}

\usepackage[utf8]{inputenc} 
\usepackage[T1]{fontenc}    
\usepackage{url}            
\usepackage{booktabs}       
\usepackage{amsfonts}       
\usepackage{nicefrac}       
\usepackage{microtype}      
\usepackage[svgnames]{xcolor}         
\usepackage{wrapfig}

\usepackage{algorithm}
\usepackage{algpseudocode}

\usepackage{caption}
\usepackage{subcaption}
\usepackage[colorlinks=true, linkcolor=Navy, citecolor=Navy]{hyperref}

\usepackage{amsmath,amsfonts,bm}

\usepackage{amsmath}
\allowdisplaybreaks 
\usepackage{amssymb}
\usepackage{mathtools}
\usepackage{amsthm}

\theoremstyle{plain}
\newtheorem{theorem}{Theorem}[section]

\theoremstyle{definition}
\newtheorem{definition}[theorem]{Definition}

\theoremstyle{remark}

\newtheoremstyle{named}{}{}{\itshape}{}{\bfseries}{.}{.5em}{\thmnote{#3}}
\theoremstyle{named}

\newcommand{\alg}{\mathcal{A}}
\newcommand{\unlearnalgo}{\mathcal{U}}

\newcommand{\dataset}{\mathcal{D}}
\newcommand{\forgetset}{\mathcal{S}}
\newcommand{\retainset}{\mathcal{D} \setminus \mathcal{S}}

\newcommand{\defn}[1]{\emph{#1}}

\usepackage{xspace}

\newcommand{\untraining}{\textsc{Untraining}\xspace}
\newcommand{\unlearning}{\textsc{Unlearning}\xspace}

\newcommand{\untrained}{\textsc{Untrained}\xspace}

\newcommand{\unlearn}{\textsc{Unlearn}\xspace}

\newcommand{\training}{\textbf{Training}\xspace}
\newcommand{\learning}{\textbf{Learning}\xspace}

\title{We Should Distinguish Unlearning From Untraining}

\author{%
  Eleni Triantafillou \\ Google DeepMind
  \And
   Ahmed Imtiaz Humayun \\ Google Research
   \And
   M\'onica Ribero \\ Google Research
  \AND
   Alexander Matt Turner
   \\ FAR AI 
   \And
   Michael Curtis Mozer \\ Google DeepMind
   \And 
   Georgios Kaissis \\ Hasso Plattner Institute
}

\begin{document}

\maketitle

\begin{abstract}
There has been a recent surge in interest about the question of how we can ``delete'' specific data points or behaviours from a trained model, a goal referred to as ``machine unlearning''. We argue that the umbrella term ``unlearning'' actually spans two distinct problem formulations, but the distinction between them has not yet been observed in literature. This causes ambiguity around when an unlearning algorithm is expected to work, leads to the use of inappropriate metrics and baselines when comparing algorithms to one another, difficulty in interpreting results, and missed opportunities for pursuing critical research directions.    
In this paper, we argue for the position that a fundamental distinction must be made between two notions that we refer to as \unlearning and \untraining. 
On one hand, \untraining aims to reverse the effect of having trained on a given ``forget set'', i.e. to remove the influence that that specific set of examples had on the model during training. On the other hand, \unlearning aims not just to remove the influence of those given examples, but also of the entire underlying subdistribution from which they were sampled (and thus e.g. the concept or model behaviour that those examples represent). 
We discuss technical definitions of these problems and map problem settings studied in the literature to each notion. 
By disambiguating technical definitions, our work aims to accelerate progress in this important field.
\end{abstract}

\section{Introduction}
``Unlearning'' was first coined by \citet{cao2015towards}, who envisioned systems that are ``capable of forgetting certain data and their lineages, completely and quickly''. 
Most early work on ``unlearning'' was motivated by regulations such as EU's General Data Protection Regulation \citep{mantelero2013eu} that stipulate that individuals have the ``right to be forgotten''. 
Specifically, in the event where an individual exercises this right \textit{after} their data has already been used to train a machine learning model, we require technical solutions for ``deleting'' that data from the model, and research in ``unlearning'' aims to address this need \citep{neel2021descent,sekhari2021remember,golatkar2020eternal,bourtoule2021machine,golatkar2020forgetting,thudi2022unrolling}. 

However, more recently, several works propose methods that use ``unlearning'' for a wider range of use cases, including removing distributions e.g. toxic language or biases \citep{allouah2026distributional}, dangerous ``knowledge'' that could e.g. aid a malicious actor to develop biological, cyber-, and chemical weapons \citep{li2024wmdp}, removing harmful ``capabilities'' or ``concepts'' to make models safer \citep{liu2024towards,yao2024large,lucki2024adversarial,barez2025open,zhang2024forget,lynch2024eight,fan2023salun,liu2025rethinking}, erasing backdoors \citep{liu2022backdoor}, eliminating poisoning attacks \citep{schoepf2024potion,schoepf2025redirection,pawelczyk2024machine}, forgetting copyrighted content like the ``Harry Potter'' books \citep{eldan2023s,shi2024muse} or specific artistic styles \citep{zhang2024unlearncanvas,fan2023salun}. All of these use cases represent important, timely and challenging problems that are only exacerbated by the increasingly large scale of modern models and their training sets, and it is therefore crucial to make progress in these directions.
However, we argue that nesting all  these use cases under the term ``unlearning'' is problematic.
Specifically, we argue for the position that \textbf{research efforts that are referred to as ``unlearning'' span two distinct problem formulations, and the lack of recognition of this entanglement causes ambiguity of objectives, false expectations about when an algorithm is expected to work,  use of inappropriate baselines or evaluation metrics, difficulty in clear interpretation of empirical results, and missed opportunities for identifying and pursuing critical research questions in the important area of post-processing models to remove the influence of data, knowledge, or behaviours}. 

In this paper, we begin to address this issue by making a novel distinction between two notions that we refer to as \unlearning and \untraining, two distinct problem formulations that, while individually well-studied, are both unfortunately referred to as ``unlearning''. 
In short, the key distinction is that \untraining simply removes the effect of having trained on the forget set, i.e. the influence of the specific forget set examples. On the other hand, \unlearning goes beyond removing (the influence of) those specific examples and generalizes to removing the influence of the entire underlying sub-distribution from which those examples are sampled (e.g. the concept that those examples represent). Notable examples from the literature are ``unlearning'' user data for privacy purposes, which is actually an \untraining problem, vs. removing dangerous model behaviours using a set of specific examples of those behaviours, which is an \unlearning problem.

In the following sections, we will first establish important terminology relating to training, learning and example influence and review the formal definition of ``unlearning'' presented in prior work. We will then draw parallels between learning and training towards the two notions that we refer to as \unlearning and \untraining, respectively. 
We sketch a technical definition for \unlearning and contrast it against the established definition of \untraining. We also map each of these technical definitions to problem settings studied in the literature, discuss the practical implications of the entanglement of \unlearning and \untraining, discuss the call for action stemming from our position and engage with alternative views.






\section{Background: Training, learning and example influence}

To set the scene for discussing our proposed distinction between \untraining and \unlearning, we devote this section to establishing important terminology and revisiting the definitions of training and learning, and related concepts of memorization, generalization and example influence.

\training, from a statistical learning theory perspective \citep{vapnik2013nature}, can be defined as the process of obtaining a function that minimizes \textbf{empirical risk on a finite dataset} $\mathcal{D}$, which generally involves solving an optimization problem. For parametric functions like neural networks, the goal of training algorithm $\mathcal{A}$ is to find a set of parameters $\theta$ that minimizes the error on a specific finite set of data $\mathcal{D}$.


\learning, on the other hand, is defined as reducing \textbf{expected risk on the underlying data distribution}. In machine learning, the goal of a training algorithm is to lead to learning. That is, the goal is to train on a finite set of samples $\mathcal{D}$ and \textbf{generalize} for the underlying data distribution from which $\mathcal{D}$ was sampled. In that sense, the notions of learning and generalization are closely related, and we treat them as synonymous in this paper.


\textbf{Does \training always lead to \learning?} 
In many cases it does.  But there are also cases where ``overfitting'' happens: the phenomenon where the training loss decreases (i.e. risk minimization is successful on the training data) but the loss of held-out examples increases (i.e. generalization does not occur). 
To address this, there is a vast literature exploring the design of training algorithms that optimally result in learning, e.g., optimization techniques like sharpness aware minimization \citep{foret2021sharpnessaware} that use the loss landscape geometry to reach a flatter minimum, techniques like MixUp \citep{zhang2017mixup} that improve generalization by augmenting the training data with convex combinations of samples, and regularization techniques like spectral normalization of parameters \citep{miyato2018spectral} that improves Lipschitz continuity of the learned function.

Recent work has also observed a phenomenon referred to as \textit{grokking} or delayed generalization \citep{power2022grokking,liu2023omnigrok,humayun2024grok} 
, where deep networks initially perfectly fit the training data 
but perform poorly for a held out dataset from the underlying data distribution---until a large number of training iterations. 

This discussion points to an important finding: a finite training dataset $\mathcal{D}$ may or may not influence the behavior of the obtained function for samples that are not in $\mathcal{D}$.
In the following section, we formalize how examples included in $\mathcal{D}$ influence training and learning in different ways.



\subsection{Influence of examples on training}
The way in which different examples influence the function obtained via training remains an active area of research \citep{jaeckel1972infinitesimal,koh2017understanding,pruthi2020estimating}. 
A phenomenon that has been studied extensively is the fact that certain examples of $\mathcal{D}$ can influence training very strongly, possibly causing those examples to be \textbf{memorized} by the resulting model. 
\citet{feldman2020does,zhang2023counterfactual} define different notions that capture ``counterfactual memorization''. The counterfactual memorization score for a training example according to \cite{feldman2020does} is given as follows.

\begin{definition}
\label{defn:mem}
\textbf{Memorization score \citep{feldman2020does}.} The \defn{memorization score} for an example $x_i \in \dataset$, with respect to a training dataset $\dataset$ and training algorithm $\alg$ is
\begin{equation}
\begin{split}
    \text{mem}(\alg, \dataset, i) &= \Pr_{f \sim \alg(\dataset)}[f(x_i) = y_i]  - \Pr_{f \sim \alg(\dataset \setminus \{x_i\})}[f(x_i) = y_i]
\end{split}
\end{equation}

where the probability distributions are over the randomness of the algorithm $\mathcal{A}$ and where $x_i$ and $y_i$ are the feature and label, respectively, of example with index $i$.
\end{definition}

The first term in the above equation considers models trained on all of $\dataset$ whereas the second term considers the counterfactual world of  models trained on $\dataset$ excluding example $x_i$. Intuitively, the memorization score for an example $x_i$ is high if including it in training results in a higher probability of the function producing the target label $y_i$ for $x_i$ compared to excluding it from training. \footnote{While the above definition assumes a classification problem, a more general notion of \textit{counterfactual memorization} has been proposed in \citep{zhang2023counterfactual} that can be applied to Large Language Models (LLMs) or other generative models. The arguments made in this paper hold for any type of architecture and training algorithm.}

Recent works \citep{feldman2020does,feldman2020neural,jiang2020characterizing} find that atypical points (i.e. points from a low-density region of the data-generating distribution) such as mislabeled data points, are more highly memorized, since these are examples that would not have been predicted correctly unless they were part of the training dataset. For example, in a classification task, imagine a training example that is an image of a cat being labeled as a chair. During training, the model can fit this strange data point and predict its assigned label of chair. However, had this example been excluded from the training set, the model would not predict that this cat is a chair. This discrepancy between the prediction on this data point of models that include it in training, compared to models that exclude it, leads to this data point being highly memorized according to Definition \ref{defn:mem}. 

Generally, the interaction between memorization and learning is an important active area of research, with theory works advocating that some memorization is even necessary for learning \citep{feldman2020does,attias2024information}. 
However, while memorization is sometimes desirable, it is also in some cases unwanted, as it may cause vulnerability to membership inference attacks or data extraction attacks \citep{shokri2017membership,carlini2022membership,carlini2021extracting}.

\paragraph{From self-influence to cross-influence.} 
Memorization according to Definition \ref{defn:mem} can be seen as a form of ``self-influence'' \citep{kulynych2025unifying}, since it measures, in expectation, how much the predictions of an example would change when including or excluding \textit{that same example} from the training set. In other words, the counterfactual memorization score is a measure of an example's influence on itself. 
\citet{kulynych2025unifying} also define a notion of \textit{cross-influence} as follows.

\begin{definition}
\label{defn:xinf}
\textbf{Cross-influence \citep{kulynych2025unifying}.} The \defn{cross-influence} of an example $x_i \in \dataset$ towards another example $x_j \in \dataset$ with $i \neq j$, with respect to a training dataset $\dataset$ and training algorithm $\alg$ is
\begin{equation}
\begin{split}
    \text{xinf}(\alg, \dataset, i, j) &= \Pr_{f \sim \alg(\dataset)}[f(x_j) = y_j] - \Pr_{f \sim \alg(\dataset \setminus \{x_i\})}[f(x_j) = y_j]
\end{split}
\end{equation}

where as before, the probability distributions are over the randomness of the algorithm $\mathcal{A}$. 
\end{definition}

Intuitively, the cross-influence of $i$ towards $j$ measures the influence that including or excluding example $x_i$ from the training data has on the predictions for example $x_j$.



\section{The formal ``unlearning'' definition}
Before introducing the distinction between \unlearning and \untraining, we first describe the classic problem formulation of ``unlearning'', which is motivated by the need to remove the influence of specific examples from a model, e.g. to enable individuals to exercise their right to be forgotten.

Let $\alg(\dataset)$ denote the weights of a model obtained by applying learning algorithm $\alg$ on dataset $\dataset$; we refer to this as the ``original model''. Informally, according to the definition of machine unlearning established in prior work, the goal of an ``unlearning'' algorithm is to remove the influence of a forget set $\forgetset \subset \dataset$ from the weights of the original model. A straightforward solution for this problem is to simply retrain a model from scratch on an adjusted training set that excludes $\forgetset$, referred to as the ``retain set'', i.e. to compute $\alg(\retainset)$. 
However, retraining from scratch is inefficient, especially for larger models. In response, ``approximate unlearning'' attempts to post-process the original model, rather than throwing it away, with the aim being to design an algorithm $\unlearnalgo$ that takes $\alg(\dataset)$ and $\forgetset$ to produce an unlearned model  $\unlearnalgo(\alg(\dataset), \forgetset, \dataset)$ that approximates the ideal solution of having trained from scratch without $\forgetset$, but arrives at that approximation much more efficiently.

Variations of technical definitions have been proposed that formalize this intuition \citep{sekhari2021remember, gupta2021adaptive, neel2021descent}, drawing inspiration from differential privacy \citep{dwork2006differential}.

\begin{definition}{\bf $(\varepsilon, \delta)$-unlearning \citep{neel2021descent}.} 
\label{defn:classic_unlearning}
For a fixed randomized learning algorithm $\alg$, an unlearning algorithm $\unlearnalgo$ is $(\varepsilon,\delta)$-unlearning with respect to $\alg$ if for any dataset $\dataset$, forget set $\forgetset \subset \dataset$, it holds that for all $R \subseteq \mathcal{R}$ we have:
\begin{align*}
    &\Pr[\textcolor{teal}{\alg(\retainset)} \in R] \le e^\varepsilon \Pr[\textcolor{violet}{\unlearnalgo(\alg(\dataset), \forgetset, \dataset)} \in R] + \delta,
    &\text{and} \\
    &\Pr[\textcolor{violet}{\unlearnalgo(\alg(\dataset), \forgetset, \dataset)} \in R] \le e^\varepsilon \Pr[\textcolor{teal}{\alg(\retainset)} \in R] + \delta.     
\end{align*}
where $\mathcal{R}$ denotes the output space, in this case, the space of model parameters.
\end{definition}

\begin{figure*}[t]
    \centering
    \includegraphics[width=0.9\textwidth]{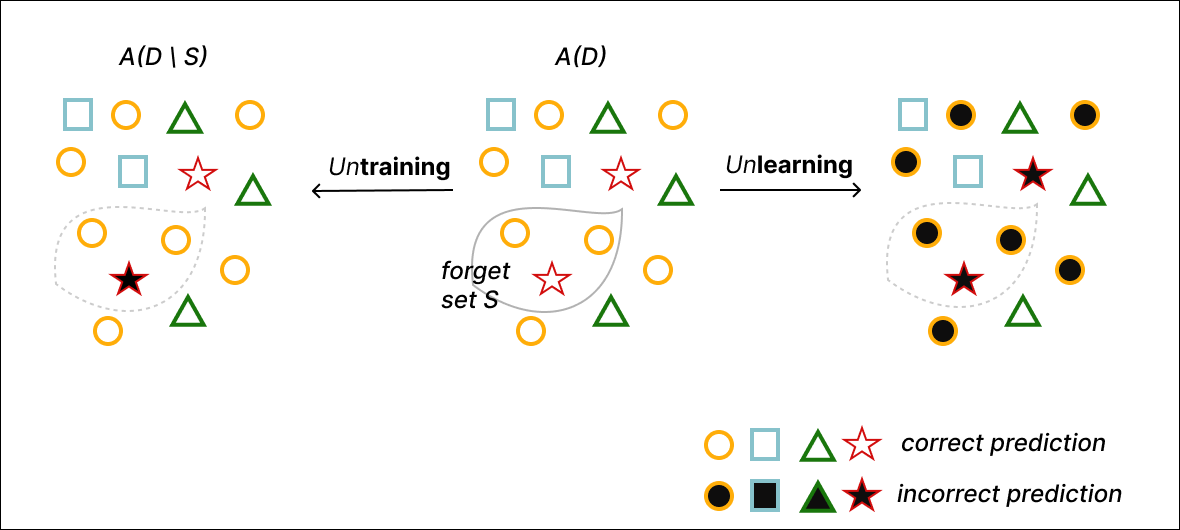}
    \caption{
    \textbf{Illustration of the difference between \unlearning and \untraining on a forget set $\forgetset$.}
    The different shapes correspond to data points in a dataset $\dataset$. A shape is white if the model predicts it correctly and black otherwise.
    In the middle, we depict the predictions of the original model that was trained on all of $\dataset$, before any \unlearning or \untraining is carried out. This model predicts correctly on all of its training set.
    On the left, we show the predictions of the model obtained by perfectly \untraining $\forgetset$, which matches the behaviour of a model trained on $\dataset \setminus \forgetset$  -- i.e. a model trained on all of $\dataset$ except for the star and the two circles that belong to the forget set $\forgetset$. 
    Notice that the \untrained model still predicts the circles correctly. This is because there are several other circles in the remaining dataset from where the model trained on $\dataset \setminus \forgetset$ can learn about circles. On the other hand, the star that was in the forget set is no longer predicted correctly, as there are fewer other stars in the dataset.
    On the right, we illustrate \unlearning of the ``behaviour'' underlying $\forgetset$, which in this case is ``being a circle or a star''. For this illustration, we assume that retraining without any examples of that behaviour  results in a model that makes incorrect predictions on all examples of that behaviour but correct predictions otherwise; the model resulting from \unlearning matches this reference.
}
\label{fig:figure1}
\end{figure*}

Measuring success of unlearning according to this definition requires estimating how close two distributions are to one another: the distribution of unlearning, i.e. $\textcolor{violet}{\unlearnalgo(\alg(\dataset), \forgetset, \dataset)}$, and that of retraining from scratch, i.e. $\textcolor{teal}{\alg(\retainset)}$.  We refer to distributions here since running each of these two procedures with different random seeds that control, for instance, the initialization and order of mini-batches, will yield slightly different model weights each time. \citet{triantafillou2024we,hayes2024inexact,kurmanji2024towards,pawelczyk2023context} discuss this issue of evaluation in more depth and propose rigorous evaluation procedures for this definition.

Note that successful ``unlearning'', according to this definition, is not (always) associated with misclassifying the examples of the forget set: if the retrained model predicts an example correctly, so should the ``unlearned'' model. And the retrained model may predict some of the examples of the forget set correctly, even though they were not in its training data, because of generalization. In other words, ``unlearning'', according to this definition, is not about removing all knowledge about the forget set; it's about removing only the \textit{additional} knowledge about the forget set that existed only \textit{due to having trained on the forget set}, i.e., the counterfactual influence of the forget set. 


In the following section, we argue that this classic definition of unlearning is better described as \untraining. We will contrast it with a different problem that we will refer to as \unlearning.

\section{Distinguishing \unlearning from \untraining} \label{sec:untr-vs-unl}

In this section, we present our proposed distinction between the two notions of \untraining and \unlearning, both of which are referred to as ``unlearning'' in the literature.


\paragraph{\untraining} As training is about minimizing empirical risk on a finite dataset $\dataset$, \untraining on $\forgetset \subset \dataset$ is the process of reversing the empirical risk minimization on $\forgetset$. In other words, the ideal solution to \untraining is to find the model parameters minimizing empirical risk on only $\retainset$. This is aligned with Definition \ref{defn:classic_unlearning}.

\paragraph{\unlearning} On the other hand, since learning is about generalization, the term \unlearning should be reserved for a different problem setting, that also shares an element of generalization. 
We propose that \unlearning a pattern, ``concept'' or ``behaviour'' from a given (representative) forget set $\forgetset$ should describe the goal of generalizing the removal beyond the specific examples in $\forgetset$, to remove the entire underlying conditional distribution representing that pattern, concept or behaviour. In this case, the goal is to approximate a model that was never trained on \textit{any} instance of that sub-distribution. 
Learning and \unlearning then both describe goals that are tied to generalization: learning induces generalization beyond the given training set $\mathcal{D}$, and \unlearning generalizes beyond the given forget set $\forgetset$, to remove the influence of the underlying sub-distribution that $\forgetset$ belongs to.

\subsection{Technical formulations}

\paragraph{``Unlearning'' of Definition \ref{defn:classic_unlearning} is actually \untraining.} Definition \ref{defn:classic_unlearning} clearly maps on to the notion of \untraining that we have described where the goal is to produce a model that is indistinguishable (in distribution) from one retrained from scratch on the retain set $\mathcal{D} \setminus \mathcal{S}$, therefore removing the influence of only the specific examples in $\forgetset$ from the model which can be seen as reversing the empirical risk minimization on $\forgetset$.

\paragraph{Defining \unlearning}
We present a sketch of a definition for \unlearning, and then contrast it with \untraining below.


\begin{definition}{\unlearning.} 
\label{defn:actual_unlearning} For a dataset $\dataset$, let $\forgetset^{\text{full}} \subset \dataset$ be the complete set of examples of $\dataset$ that capture a behaviour $\mathcal{B}$. Let $\forgetset \subseteq \forgetset^{\text{full}}$ denote the forget set that contains some representative examples of $\mathcal{B}$. 
Then, for a fixed randomized learning algorithm $\alg$, an \unlearning algorithm $\unlearnalgo$ is one such that 
$\textcolor{violet}{\unlearnalgo(\alg(\dataset), \forgetset, \dataset)}$ is indistinguishable (in distribution) from  $\textcolor{brown}{\alg(\dataset \setminus \forgetset^{\text{full}})}$. 
\end{definition}


We make the following remarks.
\begin{enumerate}
     \item We use the terms ``behaviour'' and ``concept'' interchangeably. 
     We can similarly apply \unlearning for a  ``concept'', where $\forgetset^{\text{full}}$ is then the set of all examples of that concept. 
    \item Indistinguishability can be defined via an estimate of closeness between distributions e.g. as in 
  Definition \ref{defn:classic_unlearning}, or through statistical divergences. We purposefully keep this abstract in this paper and leave it to future work to explore different instantiations, analogously to how it has been done for \untraining and Differential Privacy. 
     \item The goal of unlearning the behaviour $\mathcal{B}$ using the forget set $\forgetset$ would be the removal of the cross influence of $\forgetset^{\text{full}}$ on all examples in $\dataset$.
    \item This is a conceptual definition to illustrate the notion of \unlearning (contrasting it to the notion of \untraining), but in practice we may not be able to specify the set $\forgetset^{\text{full}}$ containing every instantiation of a behaviour in a given dataset (when $\forgetset$ is a strict subset of $\forgetset^{\text{full}}$, it doesn't uniquely specify $\mathcal{B}$).
    \item When $\forgetset$ is very small, relative to $\forgetset^{\text{full}}$, this may be referred to as ``few-shot'' \unlearning \citep{yoon2024few,de2024unlearning}.
    \item If $\forgetset$ includes all training instances of $\mathcal{B}$, then \untraining $\forgetset$ and \unlearning $\mathcal{B}$ are one and the same. \footnote{The only other case where \untraining $\forgetset$ and \unlearning $\mathcal{B}$ are the same is where the data manifold is such that removing $\forgetset$ destroys the complete manifold of $S^{\text{full}}$, but this case will not happen in practice.}
\end{enumerate}

Notice that the definition of \unlearning has an element of generalization: from the specific forget set $\forgetset$ showcasing a ``behaviour'' $\mathcal{B}$, \unlearning \textit{generalizes} to removing the entire sub-distribution of $\mathcal{B}$, beyond $\forgetset$; analogously to how learning from a specific set of examples also leads to generalization in the sense of acquiring broader knowledge beyond the given set of examples. This is the key aspect that distinguishes \unlearning from \untraining. And this generalization of knowledge removal is crucial for removing unwanted behaviours from models, as in practice it's not possible to specify every possible instantiation of a behaviour $\mathcal{B}$ in $\mathcal{S}$.

\paragraph{\untraining and \unlearning are materially distinct.} 
Notice that \untraining a forget set $\forgetset$ does not mean that the model is unable to predict the examples of $\forgetset$ correctly. A model trained purely on $\retainset$ may still be able to predict the examples of $\forgetset$ correctly due to generalization.  Specifically, \untraining a forget set $\mathcal{S}$ will lead to:
(i) No change to the model, if $\mathcal{S}$ has low self-influence and low cross-influence; 
(ii) Making the model unable to predict on $\mathcal{S}$ but without harming the model's predictions on any other examples, if $\mathcal{S}$ has high self-influence but low cross-influence;
(iii) Making the model unable to predict on $\mathcal{S}$ while also harming the model's predictions on other examples, if $\mathcal{S}$ has high self-influence and high cross-influence. On the other hand, \unlearning a pattern, concept or ``behaviour'' from a forget set $\forgetset$ aims to always make the model unable to predict the examples of $\forgetset$ any better than a model that never trained on \textit{any} instances of the concept (i.e. a model that never learned the concept). 
We illustrate the difference between \untraining versus \unlearning, for a given forget set $\forgetset$ in Figure \ref{fig:figure1}.

\textbf{Cases where the outcomes of \untraining and \unlearning for the same forget set differ.}  
Consider the scenario where $\forgetset$ contains only non-memorized examples (examples with very low memorization scores) of a ``behaviour'' $\mathcal{B}$. In that case, the solution to \untraining would be to simply leave the model as is\footnote{This observation has been previously made by \citet{zhao2024makes}, who also studies how existing unlearning algorithms perform on forget sets of different degrees of memorization.}, since that approximates well the desired reference point of \untraining, namely $\textcolor{teal}{\alg(\retainset)}$. 
On the other hand, the solution to \unlearning would be quite different, as a large modification would be required in this case to approximate a model $\textcolor{brown}{\alg(\dataset \setminus \forgetset^{\text{full}})}$ that never trained on \textit{any} example of the behaviour that $\forgetset$ represents.

To give a different example, a model that was trained with Differential Privacy (DP) \citep{dwork2006differential} is one that does not memorize any of its training examples (more than permissible). Intuitively, for a DP model, \untraining a single example is then a no-op (though \untraining a larger number of examples requires group-privacy accounting that weakens the privacy guarantee\footnote{More precisely: a training algorithm satisfying $(\varepsilon,\delta)$-DP under add/remove adjacency, automatically satisfies the corresponding approximate \untraining requirement for a single-record deletion. Larger forget sets require group-privacy accounting: even for pure DP, the standard bound becomes (k$\varepsilon$) for (k) records. }), whereas \unlearning is not, and would in fact require substantial modification to take the model from a state where it generalizes perfectly to one that has no knowledge about the sub-distribution relating to the behaviour that it is asked to \unlearn. \footnote{In fact a relevant discussion was held in the DP field, see e.g., {\url{https://github.com/frankmcsherry/blog/blob/master/posts/2016-08-16.md}}}

\section{Why does establishing this distinction matter?} 
So far, we have argued that the problem settings of \unlearning and \untraining are distinct. We argued this both through distinguishing their technical definitions and through reasoning about how their outcomes are materially different in different cases. We now further argue that the distinction between \unlearning and \untraining is not just of academic interest, but has important practical implications: lacking this distinction can lead to several issues, summarized below.

\textbf{Unclear expectations. } The lack of clear specification of problem settings can leave us with unclear expectations for when an algorithm should work well. 
There are recent examples in the literature effectively showing that algorithms designed for \untraining perform very poorly for what we would refer to as \unlearning problems, but without recognizing why that is, or identifying the distinction between those problem settings that we establish in this paper.   
As an example, 
\citet{goel2024corrective,schoepf2025redirection} study ``unlearning'' corrupted data, in the settings of full discovery (all corrupted data is known and is placed in the forget set) and partial discovery (only a subset of the corrupted data is known and placed in the forget set). 
\citet{goel2024corrective} show that algorithms developed for \untraining (in their terminology, the ``full discovery'' setting) fail catastrophically in the partial-discovery setting, where only a subset of the corrupted data is discovered; a setting requiring an \unlearning algorithm that can \textit{generalize}. This highlights that the two problem settings of \untraining and \unlearning are fundamentally distinct, and we may falsely expect an algorithm that performs well for one to also work well for the other, if we lack a clear distinction between these settings. 
\citet{schoepf2025redirection} initiates the study of how the ``statistical regularity'' of the ``corruption'' (or, in our more general case, the ``concept'' or the ``behaviour'' that we want to \unlearn) affects how successfully different algorithms can generalize to remove the entire ``corruption'' under partial discovery.

\textbf{Ambiguity around evaluation metrics. } 
Authors of ``unlearning'' papers may choose to use inappropriate metrics and baselines in their evaluation, and reviewers of papers may request the inclusion of inappropriate metrics and baselines, further adding to the confusion. 
Many ``unlearning'' metrics have been proposed, corresponding to different threat models (e.g white-box vs black-box access to the ``unlearned model) and different use cases \citep{hayes2024inexact,golatkar2020eternal,kurmanji2024towards,triantafillou2024we,lucki2024adversarial,shi2024muse,maini2024tofu}. Not all of these metrics are appropriate for both \untraining and \unlearning, so this distinction is necessary for ensuring that we evaluate in a reliable way that is informative of an algorithm's performance for its intended use case. For example, if the goal is to \unlearn a dangerous behaviour for safety, then using a Membership Inference Attack (MIA) \citep{shokri2017membership,carlini2021extracting} to attempt to infer whether the examples of the forget set were part of the training set of the model, is not appropriate, despite this being an established metric for \untraining. The reason that this is not appropriate for \unlearning is that the goal of unlearning is not to remove the influence of \textit{only} the forget set examples: it might be the case that the MIA fails to identify that any of the forget set examples were once included in the model's training set (thus passing the empirical test of \untraining), but the model continues to have knowledge of the sub-distribution that the forget set was sampled from (unsuccessful \unlearning), as would have been evidenced, for instance, by the success of an MIA to identify that \textit{other} (non-forget set) examples from that distribution were in fact part of the training dataset.
As another example, out of the evaluation metrics of the MUSE benchmark \citep{shi2024muse}, do we always require \textit{every} ``unlearning'' algorithm to satisfy \textit{all} the criteria, e.g. ``no verbatim memorization'' and ``no knowledge memorization'' and ``no privacy leakage''? We again argue that it depends on the underlying problem specification.

\textbf{Difficulty in interpreting results and transferring insights. } 
\citet{li2025llm} recently observed an interesting behaviour that occurs with gradient ascent and the gradient ascent-based NPO method \citep{zhang2024negative}. 
Specifically, they find that, when using these methods, the resulting ``unlearned'' model indeed no longer generates the target sequence it was asked to forget, but is still able to generate semantically-related outputs that still contain the ``unwanted knowledge''.
This is because applying these methods on specific target responses leads to redistributing the probability mass to other tokens and sequences that may actually correspond to semantically-related paraphrasing of the targets that should have been forgotten. 
But is this really a failure mode of NPO? Is NPO designed to be an \untraining or an \unlearning algorithm? Should we expect it to excel at both? These are open research questions that have not received sufficient attention \textit{because} this important distinction had not been previously established. 
To put it in our terminology, \citet{li2025llm} are interested in \unlearning the unwanted behaviour, from a given specific instantiation of it, not \untraining that specific instantiation only. Indeed, their proposed solution is to modify the algorithm to suppress not only specific target responses, but also ``the model's beliefs'', effectively turning an \untraining algorithm into an \unlearning one. 
Establishing a common language that describes these issues will allow to better share insights and interpretations of observed phenomena, and proposed solutions, across individual research papers. 

\textbf{Critical research questions are left unexplored.} 
The lack of a distinction between \unlearning and \untraining has led to leaving important research questions unexplored, which we hope that future work pursues, including: (i) Which existing ``unlearning'' algorithms are better suited for \untraining compared to \unlearning? (ii) How large does $\forgetset$ need to be relative to $\mathcal{S}^{\text{full}}$ for unlearning to succeed? In what ways does this depend on the \unlearning algorithm, the ``representativeness'' of $\mathcal{S}$ and the characteristics (e.g. statistical regularity)  of the behaviour $\mathcal{B}$?, (iii) How data efficient are different \unlearning algorithms and what are their trade-offs between data-efficiency vs precision? (iv) Can we characterize when we are better off spending more resources to find a larger subset of $\mathcal{S}^{\text{full}}$ vs relying on the inherent generalization power of the \unlearning algorithm? (v) Most ``unlearning'' algorithms were traditionally developed for \untraining. Can we now also create algorithms that are explicitly designed with the appropriate inductive biases for generalization beyond the forget set in mind, perhaps drawing inspiration from algorithms developed for \textit{learning} from few examples?

\section{Mapping \untraining and \unlearning to the literature}

\textbf{Examples of \untraining.} 
The most prominent example, which was a key motivation behind several ``unlearning'' (more accurately, \untraining) works, is that of protecting user privacy, by honoring users' requests to ``delete'' their data from models \citep{neel2021descent,sekhari2021remember,golatkar2020eternal,bourtoule2021machine,golatkar2020forgetting,thudi2022unrolling}.
Another example is removing (the training influence of) data points that are mislabeled, outdated, or noisy, in order to correct their adverse effects on the model's predictions  \citep{kurmanji2024towards,goel2022towards}. In generative models, several recent works have studied removing the training influence of specified copyrighted texts or other extractable records, such as copyrighted sentences from the Harry Potter books \citep{eldan2023s} or specific data points from LLMs   \citep{jang2023knowledge,barbulescu2024each}, or diffusion models \citep{alberti2025data}, aiming to reduce the extractability of these records, motivated by privacy or copyright issues. 


\textbf{Examples of \unlearning.} 
There are several recent examples of \unlearning in the literature, including \unlearning dangerous knowledge or capabilities \citep{li2024wmdp,liu2024towards,lynch2024eight} \footnote{Recent research in LLMs is conducted on top of large pretrained models, where we don't have control of the training set and we don't have knowledge of which training examples gave rise to different ``behaviours''. It's possible that the forget sets used aren't even part of the training set. One could broaden Definition \ref{defn:actual_unlearning} to consider $\forgetset$ that isn't necessarily part of $\mathcal{D}$.}, erasing backdoors \citep{liu2022backdoor}, or concepts like ``not safe for work'' \citep{fan2023salun,zhang2024forget}, where we have a representative set of examples representing each of these behaviours or concepts, which is a strict subset of the overall set of such examples. 
The notion of ``corrective unlearning'' \citep{goel2024corrective} is also a type of \unlearning: the goal is to remove a ``corruption'' or poison from partial discovery of the training data that cause the corruption or poison. This topic is enjoying increasing attention recently \citep{schoepf2024potion,schoepf2025redirection}. Similarly, \unlearning an ``artistic style'' from some examples of that style falls into this category \citep{fan2023salun,zhang2024unlearncanvas}. 
Finally, \citet{zhang2024theft} recently observed a phenomenon they refer to as ``ripple effect of unlearning'', which refers to the ``generalization'' of unlearning algorithms to unlearn several ``harms'', even when those aren't explicitly present in the forget set. They give an example where a model instructed to ``unlearn'' the steps for theft may also implicitly ``unlearn''
the steps for making a bomb. They hypothesize that this is possible due to the intrinsic relatedness of harmful responses, across different types of harms. In our terminology, this phenomenon can be seen as \unlearning the ability to respond in a harmful manner, from some representative examples. 


\section{Alternative views}

\paragraph{Alternative view: The distinction between \unlearning and \untraining already exists: ``data unlearning'' versus ``class/concept unlearning''.}

It might appear that these two distinctions are related, but they are in fact distinct. 
As we discussed at length, both the removal of individual data points as well as the removal of concepts can be operationalized through ``unlearning'' specific data points, i.e. through feeding a specific forget set (``data'') to an ``unlearning algorithm''. But this procedure might cause several different outcomes, ranging from (i) not having any effect, to (ii) removing the influence of a given data point from the model but not affecting anything else (e.g if this data point has low cross-influence), to (iii) removing an entire underlying concept. See Section \ref{sec:untr-vs-unl}.  
Should the term ``data unlearning'' be used in all of these cases?
In the first case where there is no effect, would we still say that we ``unlearned'' that data? In the latter case where an entire concept is removed, would we still call that ``data unlearning'' or would we now call it ``concept unlearning''?
It's unclear if these terms refer to the \textit{outcome} or the \textit{mechanism}, and because these notions have not been formalized, they may be used loosely to refer to either, which causes confusion.

We define \untraining and \unlearning as mechanisms both of which operate on a given forget set $\forgetset$ but that achieve well-specified \textit{outcomes} that are defined as ``matching'' the reference points $\textcolor{teal}{\alg(\retainset)}$ and $\textcolor{brown}{\alg(\dataset \setminus \forgetset^{\text{full}})}$, respectively. These definitions clearly specify the relationship between the input data, and the desired outcome. Note that satisfying the definition of \untraining might still cause different outcomes \textit{in terms of the predictions made on different examples in $\forgetset$} as discussed, and \untraining may sometimes lead to \unlearning, just like training may (and often does) lead to learning. But unlike the distinction between ``data unlearning'' and ``concept unlearning'', these definitions are precise, can be mapped onto important use cases in the literature and we conjecture can be used as a solid basis for disambiguating goals, developing appropriate evaluation protocols for each and ultimately designing better algorithms for both of these important problems.

\paragraph{Alternative view: We don't need to make this distinction, as long as each individual research paper clearly states the problem setting that it studies.} It is certainly a large undertaking to coordinate definition and problem formulations across a large and rapidly growing body of literature. However, we argue that succeeding in this aspect would significantly accelerate progress through sharing insights, identifying and targeting important open questions, leveraging the most suitable algorithms and metrics for experimental investigations and ultimately removing friction towards developing better methods for each of the important problem settings in this important space.

\section{Related notions}

\paragraph{Distributional unlearning.} \citet{allouah2026distributional} proposed the notion of ``distributional unlearning'' that is conceptually similar to our definition of \unlearning in the sense that it requires generalization beyond removing the influence of specific data points. To formalize distributional unlearning, they consider two probability distributions $p_1$ and $p_2$, that model the domains that we want to remove and preserve, respectively. Then, they formalize their goal of ``distributional unlearning'' as constructing a new data distribution that is statistically distant from $p_1$ while remaining close to $p_2$. We instead define our notion of \unlearning via the reference point of retraining from scratch without $\forgetset^{\text{full}}$, illustrating clearly the difference between Definition \ref{defn:actual_unlearning} and Definition \ref{defn:classic_unlearning}. 

\paragraph{Model editing.} Model editing \citep{meng2022locating,meng2022mass,mitchell2021fast} aims to post-process a trained model in order to update or insert specific factual knowledge or behaviours, while otherwise leaving the model intact. For a given ``edit'', aside from desiring efficacy and specificity (i.e. the edited model makes the correctly-altered prediction on the examples it was instructed to without distorting its knowledge of unrelated facts or examples), another key desideratum is \textit{generalization}: the edit should hold not only for the examples whose outputs the model editing method was instructed to update, but also for examples in the ``equivalence neighbourhood" that contains related input/output pairs. For example, if we want to edit a model to alter ``The Eiffel Tower is located in the city of Paris'' to ``The Eiffel Tower is located in the city of Rome'', then we also want it to complete the sentence ``The landmark known as the Eiffel Tower can be found in '' with the word ``Rome'' as well. In that sense, model editing has a similar flavour to our notion of \unlearning: they both necessitate generalization beyond the specific set of examples that are available for performing the intervention. 
There are of course other differences between these two problems. An important difference discussed in \citep{liu2025rethinking} is that unlearning seeks to remove information while model editing aims to replace a fact or behaviour with a different one. Another difference can be seen from Equation 1 in \citep{mitchell2021fast}: model editing is concerned with the model ``behaving'' according to the desired edit (i.e. produce outputs that match the expected post-editing outputs). On the other hand, \untraining definitions from previous work and \unlearning as we define it in this paper instead require satisfying a statistical relationship between the weights of the resulting model compared to the weights of an appropriate reference point (in model editing, that reference could be the model that would have resulted from training with the correct or desired knowledge or facts from the start). We leave it to future work to formalize different notions for model editing, but we believe that the distinction that we make between \untraining and \unlearning offers a useful foundation for discussing the points of connection and differentiation between these related fields.

\paragraph{Concept erasure.}
Concept erasure aims to remove a target concept from model representations \citep{belrose2023leace} or prevent its generation \citep{gandikota2023erasing}, while preserving other information and capabilities.
In that sense, it shares with \unlearning the goal of generalising removal beyond individual training examples.
However, the success criteria differ: methods operating on representations may guarantee erasure only against a chosen class of predictors, such as linear classifiers, while generative methods typically assess outputs.
These criteria do not by themselves require matching the distribution of model parameters obtained by retraining without $\forgetset^{\text{full}}$.
Moreover, entanglement between concepts can cause erasure to affect other concepts \citep{amara2025erasing}, highlighting the importance of specifying both the target and what should be preserved.

\section{Limitations and scope}
Our goal in this paper is not a complete taxonomy of ``unlearning'' algorithms; we refer the reader to existing surveys for this \citep{nguyen2022survey}. We also don't attempt to discuss inherent limitations of attempting to remove knowledge or control capabilities post-hoc  \citep{cooper2024machine,shumailov2024ununlearning,siddiqui2026position} nor distinctions between ``deleting'' knowledge from ``suppressing'' knowledge \citep{hu2024unlearning,deeb2024unlearning,siddiqui2025dormant,che2025model,jia2025erasure}.  
We also leave out of scope the discussion of non-technical issues relating to \untraining and \unlearning but refer the reader to  \citep{cooper2024machine} for a discussion of how mismatches that exist between motivations for these problem settings and the technical methods have implications for U.S. copyright law, privacy and safety. Relatedly, \citet{yoon2026position} recently argued that we should distinguish between problems of dataset-defined deletion and policy-defined behavior control; a recommendation that stems from a similar motivation to ours, that the term ``unlearning'' is overloaded, but from a different perspective. 
We also note that our definition of \unlearning applies
only to concepts or behaviours for which removing a
well-defined subset $\forgetset^{\text{full}}$ of the
training data provides a meaningful retraining reference.
More general capabilities, such as edge detection, may
arise from many different training examples and need
not admit a natural choice of such a subset.
Removing these capabilities may therefore require
different definitions and is out of scope of this work.

\section{Conclusion and call to action}
We argued that the term ``unlearning'' has been used in the literature to describe two distinct problem formulations, that we refer to as \unlearning and \untraining. 
We established the fundamental distinction between \unlearning and \untraining, both at an intuitive level, drawing connections to the notions of training and learning, and through technical definitions. We demonstrated that these problem settings are distinct through illustrative examples and discussed why this distinction is important in practice. Our call for action for the community is that authors of papers on ``unlearning'' precisely state the problem formulation(s) that they target, whether it is a problem formulation that they propose a new algorithm for, or a new evaluation metric for. Further, we call reviewers of ``unlearning'' papers to keep in mind the problem formulation that the paper is addressing when requesting additional baselines or metrics to be included. This paper only scratches the surface of necessary work on defining the right technical subproblems in the larger space of post-processing models to alter their underlying training data, knowledge or behaviours. We hope that future work continues to push in this direction, for instance by formalizing the connection between ``unlearning'' and adjacent areas like model editing or concept erasure. We also hope that future work tackles the critical research questions that are uncovered by the distinction that we draw in this work, en route to improving the technical methods and evaluation techniques for both of these important subproblems.

\newpage

\bibliographystyle{abbrvnat}
\bibliography{neurips_2026}

\appendix





\end{document}